\documentclass[10pt,twocolumn,letterpaper]{article}

\usepackage{cvpr}              
\usepackage{multirow}
\usepackage{float}
\usepackage[accsupp]{axessibility}
\usepackage[dvipsnames]{xcolor}

\definecolor{cvprblue}{rgb}{0.21,0.49,0.74}
\usepackage[pagebackref,breaklinks,colorlinks,citecolor=cvprblue]{hyperref}
\definecolor{First}{rgb}{0.95, 0.62, 0.61}
\definecolor{Second}{rgb}{0.97,0.81,0.63}
\definecolor{Third}{rgb}{1.0, 0.97, 0.70}
\usepackage{xcolor}
\usepackage{colortbl}

\def\paperID{5962} 
\def\confName{CVPR}
\def\confYear{2024}

\title{XScale-NVS: Cross-Scale Novel View Synthesis \\ with Hash Featurized Manifold}

\author{Guangyu Wang$^1$, Jinzhi Zhang$^1$, Fan Wang$^2$, Ruqi Huang$^{1}$\footnotemark[2], Lu Fang$^{1}$\footnotemark[2]\\
Tsinghua University$^1$, Alibaba Group$^2$\\
}
\begin{document}
\maketitle
\footnotetext[2]{Lu Fang (fanglu@tsinghua.edu.cn, \href{http://www.luvision.net/}{luvision.net}) and Ruqi Huang (ruqihuang@sz.tsinghua.edu.cn, \href{https://rqhuang88.github.io/}{rqhuang88.github.io}) are the corresponding authors.}
\begin{abstract}
We propose \emph{XScale-NVS} for high-fidelity cross-scale novel view synthesis of real-world large-scale scenes. 
Existing representations based on explicit surface suffer from discretization resolution or UV distortion, while implicit volumetric representations lack scalability for large scenes due to the dispersed weight distribution and surface ambiguity. In light of the above challenges, we introduce hash featurized manifold, a novel hash-based featurization coupled with a deferred neural rendering framework. This approach fully unlocks the expressivity of the representation by explicitly concentrating the hash entries on the 2D manifold, thus effectively representing highly detailed contents independent of the discretization resolution. We also introduce a novel dataset, namely \texttt{GigaNVS}, to benchmark cross-scale, high-resolution novel view synthesis of real-world large-scale scenes. Our method significantly outperforms competing baselines on various real-world scenes, yielding an average LPIPS that is $\sim$ 40\% lower than prior state-of-the-art on the challenging \texttt{GigaNVS} benchmark. Please see our project page at: \href{https://xscalenvs.github.io/}{xscalenvs.github.io}.
\end{abstract}

\section{Introduction}
\label{sec:intro}
Neural rendering has attracted significant amount of research interests and influenced down-stream applications including virtual reality, visual effects, robotic simulation, to name a few. Among the recent advances, a notable trend {focuses} on free-view rendering of real-world large-scale scenes~\cite{liu2023real, 10274871, rematas2022urban, tancik2022block, turki2022mega, gu2023uenerfneural}. While these approaches {excel at} recovering macro-scale structures from images {captured at a distance}, they often struggle to simultaneously deliver micro-scale details,
leading to a discrepancy that hinders comprehensive visual perception of the real world (See Fig.~\ref{fig:qualitative_teaser}). {This limitation necessitates a novel scene representation tailored for high-fidelity cross-scale novel view synthesis (NVS) of real-world large-scale scenes.} In particular, we identify and address two critical challenges towards this goal as follows.

First of all, reconstructing real-world large-scale scenes in high quality requires to collect images from both distant and close-up views. However, current real-world NVS benchmarks~\cite{zhang2021gigamvs, barron2022mip, knapitsch2017tanks, schops2017multi, dai2017scannet} are limited to a single macro-scale setup that neglects close-up imagery for local details and focus primarily on small-scale scenes. To this end, we introduce \texttt{GigaNVS} dataset to benchmark cross-scale, high-resolution NVS of real-world large-scale scenes. The dataset contains seven scenes covering an average area of $1.4\times 10^6m^2$, each of which is {captured using a combination of aerial and ground photography}, 
{yielding a collection of} 1,600 $\sim$ {18,000} high-quality 5K/8K multi-view images per scene with unstructured scale variations. Remarkably, our dataset captures millimeter-level details from scenes with square-kilometer-level areas, {enabling an effective texture resolution of 30 billion during the reconstruction.} Therefore, the proposed \texttt{GigaNVS} benchmark simultaneously characterizes gigantic scene scale and unparalleled richness of real-world contents by providing cross-scale, high-resolution, and real-captured data.

Secondly, there lacks a suitable representation simultaneously possessing global robustness and local expressivity. Formulations that favor the former typically rely on an imperfect surface proxy reconstructed from multi-view stereo (MVS)~\cite{furukawa2015multi, schonberger2016pixelwise, ji2017surfacenet, yao2018mvsnet}, and then featurize the proxy on either the parametrized 2D UV map~\cite{thies2019deferred, liu2023real, 10274871} or the 3D surface~\cite{aliev2020neural, yang2022neumesh, ruckert2022adop, rakhimov2022npbg++, kopanas2022neural, zuo2022view, kerbl20233d}. While being {global}-structure-aware, such representations typically struggle to represent the local intricate details. 
For instance, as shown in Fig.~\ref{fig:qualitative_teaser} (a), UV-based featurizations~\cite{thies2019deferred, liu2023real, 10274871} suffer from the {inherent distortion} of surface parametrization~\cite{floater2005surface, Hormann2007MeshPT, Ray2006PeriodicGP}, {which severely degrades the rendering quality}.
Meanwhile, representations based on more global 3D discretization, such as point cloud~\cite{aliev2020neural, ruckert2022adop, rakhimov2022npbg++, kopanas2022neural, zuo2022view}, mesh~\cite{yang2022neumesh}, or Gaussians~\cite{kerbl20233d}, all suffer from limited featuremetric resolutions, {i.e., the resolutions of the local spatial features,} which depend on the respective discretization resolution. As shown in Fig.~\ref{fig:qualitative_teaser} (b), such feature allocation mode can not fully describe the micro details presented in real-world large-scale scenes.

On the other end of the spectrum, state-of-the-art implicit volumetric representations~\cite{li2023neuralangelo, rosu2023permutosdf, wang2023neus2, barron2023zip, muller2022instant} {offer the potential to} express arbitrarily high spatial resolutions with multi-resolution hash encoding~\cite{muller2022instant}, enabling more effective and efficient neural field reconstruction. Similar to earlier neural field based methods~\cite{mildenhall2021nerf, rematas2022urban, tancik2022block, turki2022mega, xiangli2022bungeenerf}, these approaches rely on volume rendering, which alpha-composites multiple samples along the ray, to produce effective gradient signals for a plausible geometric optimization~\cite{wang2021neus}. Unfortunately, for complex large-scale scenes, this comes at the cost of a dispersed weight distribution
and leads to surface ambiguities, since the informative surface intersection is compromised with non-informative, multi-view inconsistent samples, 
as shown in Fig.~\ref{fig:qualitative_teaser} (c). 

To address the limitations of existing scene representations, we introduce \emph{hash featurized manifold} representation, {i.e., an expressive hash-based featurization upon the surface manifold} for high-fidelity cross-scale NVS of real-world large scenes. {Our key insight is to featurize the 2D surface manifold with 3D volumetric hash encoding, sidestepping the complex surface parametrization to strictly preserve geometric conformality while leveraging rasterization to concentrate the learnable hash entries on multi-view consistent signals throughout the optimization.} Compared to existing representations based on explicit 3D discretization, our hash-based featurization {unlocks the dependence on the discretization resolution} by adaptively erasing the impact of unimportant spatial features on hash entries, and naturally bypasses any surface parametrizations to circumvent the distortions.
In turn, by explicitly prioritizing the sparse manifold instead of densely featurizing a redundant volume, the expressivity of multi-resolution hash encoding is substantially incentivized. The reason is that the extensively correlated spatial features receive clean colour gradients purely from multi-view consistent regions without any disturbance from inconsistent regions, thus better reasoning about the optimal capacity for modelling view-invariant reflectance components.

Hash featurized manifold {can be} tightly coupled with a deferred neural rendering pipeline to simultaneously advance expressivity and efficiency. To render our representation, we first obtain a screen-space feature buffer through rasterization, and then employ an Multi-Layer Perceptrons (MLPs) based neural shader to reason about the view dependent surface colour. 
We also introduce two enhancements tailored for our representation to better express the cross-scale rich contents: 
1) A surface multisampling scheme to enable a prefiltered featurization, which copes with the unstructured scale variations and eliminates aliasing by sampling the curved surface; 
2) A manifold deformation mechanism to implicitly enforce multi-view consistency  
for a better tolerance of geometric imperfections on the surface manifold. Combining the two designs together, our representation essentially {expresses} a deformable frustum near the surface rather than a single ray-surface intersection, allowing for more flexible descriptions of fine-grained details. 

The proposed method significantly outperforms prior approaches on the challenging \texttt{GigaNVS} benchmark and the
public Tanks\&Temples dataset~\cite{knapitsch2017tanks}. Remarkably, our method reduces the average LPIPS relative to the state-of-the-art by 40\% on \texttt{GigaNVS}, pushing the boundary of in-the-wild cross-scale neural rendering towards unprecedented levels of details and realism.
In summary, our main contributions are as follows:
\begin{itemize}
    \item We propose hash featurized manifold representation that fully unleashes the expressivity of volumetric hash encoding by rasterizing the surface manifold to explicitly prioritize multi-view consistency. 
    \item We present a deferred neural rendering framework to efficiently {decode} the representation and propose two tailored designs to better describe cross-scale details. 
    \item We introduce \texttt{GigaNVS} dataset to benchmark cross-scale, high-resolution NVS of real-world large-scale scenes, where our method demonstrates significant improvements over prior approaches. 
\end{itemize}



\section{Related Work}
\label{sec:related_work}

\par\noindent\textbf{Large-scale Scene NVS.} 
Recent approaches exploit spatial partitioning~\cite{turki2022mega, tancik2022block, gu2023uenerfneural} and geometric priors~\cite{rematas2022urban, 10274871, liu2023real, li2023read} to better handle large-scale scenes. However, these works can only represent scenes reasonably at a macro-scale yet exhibit excessively blurry artifacts when navigating closer to inspect micro details. Remarkably, BungeeNeRF~\cite{xiangli2022bungeenerf} enables multi-scale neural rendering of large scenes with a progressive optimization scheme to gradually expand the model and training data. However, it requires an explicit split of scales among the input images, thus being limited to remote sensing like scenarios, where the scale can be readily measured by the flight altitude. By contrast, we focus on general large-scale scenes and unstructured scale variations commonly presented in practical perception. We hold that our task can not be well addressed without modifying the fundamental scene representation.

\par\noindent\textbf{Large-scale Scene Benchmark.} 
The recent trend of 3D datasets~\cite{xiangli2022bungeenerf, li2023matrixcity, raistrick2023infinite, UrbanScene3D, yao2020blendedmvs} start to focus on large scenes, however, the available multi-view images are either synthesized by game engines~\cite{li2023matrixcity, raistrick2023infinite, UrbanScene3D} or re-rendered from reconstructed mesh~\cite{xiangli2022bungeenerf, UrbanScene3D, yao2020blendedmvs}, drastically deteriorating the diversity and fidelity of real-world contents. 
Empowered by the recent advance of gigapixel-level sensation~\cite{yuan2021modular, wang2020panda, zhang2020multiscale}, the GigaMVS dataset~\cite{zhang2021gigamvs} captures real-world large-scale scenes with ultra-high-resolution imagery. However, the collected images are shot from a distance, whose amount is also limited due to the complicated imaging procedure. To the best of our knowledge, the proposed \texttt{GigaNVS} is the first real-captured dataset targeting cross-scale, high-resolution NVS of large-scale scenes.

\par\noindent\textbf{Representations upon explicit 3D discretization.} 
Thies et al.~\cite{thies2019deferred} incorporate neural textures into traditional mesh rasterization pipeline and use a CNN-based neural renderer to enable high quality NVS.
Another line of approaches~\cite{aliev2020neural, ruckert2022adop, rakhimov2022npbg++, kopanas2022neural, zuo2022view} follow a 
similar pipeline but use point as the surface primitive and directly featurize the surface in 3D. Recently, 3D Gaussian Splatting~\cite{kerbl20233d} demonstrates great success in terms of rendering quality and efficiency with a highly flexible point-based representation. 

\par\noindent\textbf{Implicit Volumetric Representations.} The exploding neural representations~\cite{mildenhall2021nerf, barron2021mip, muller2022instant, wang2021neus, yariv2021volume, barron2022mip, li2023neuralangelo, barron2023zip} implicitly encode scene geometry as the density field~\cite{mildenhall2021nerf, barron2021mip, sun2022direct, muller2022instant, barron2022mip, barron2023zip} or signed distance field~\cite{wang2021neus, yariv2021volume, wang2023neus2, rosu2023permutosdf, li2023neuralangelo} and represent appearance as the radiance field. 
Notably, iNGP~\cite{muller2022instant} proposes multi-resolution hash encoding, which can conceptually represent a dense feature grid at arbitrarily high resolution by hashing a fixed-size learnable array. Neuralangelo~\cite{li2023neuralangelo} further extends this with a coarse-to-fine control of the hash grid and demonstrates impressive neural surface reconstruction quality. However, existing volumetric neural fields can not generalize well to large-scale scenarios due to the inherent surface ambiguities.
\begin{figure*}[htbp]
    \centering
    \newcommand{\colw}{0.19}
    \newcommand{\figw}{1} 
    \includegraphics[width=\figw\textwidth,trim={0cm 0cm 0cm 0cm},clip]{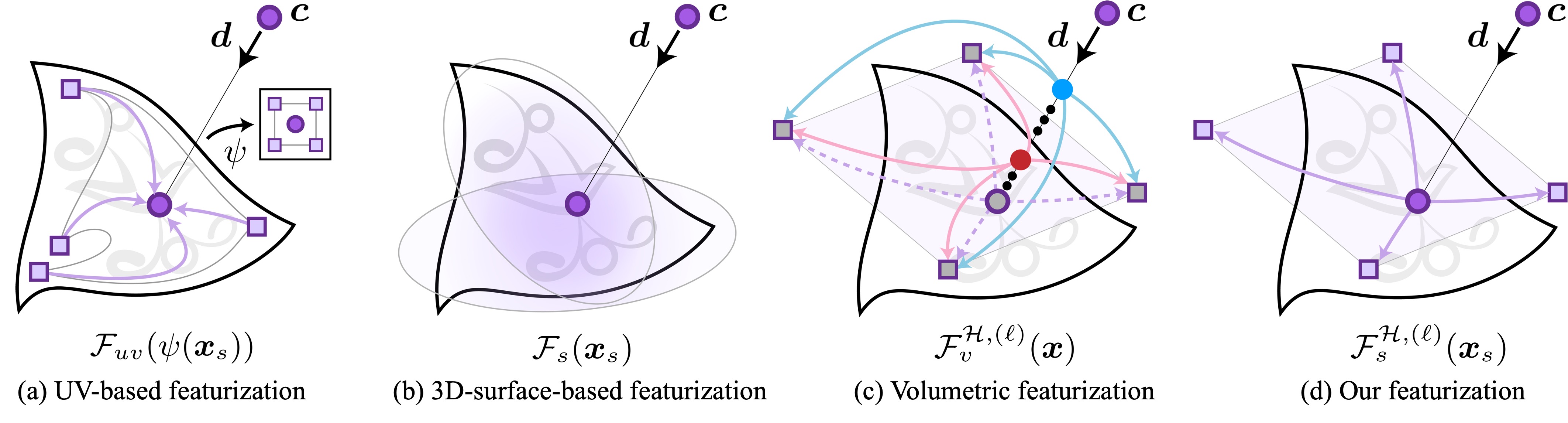}
    \hfill
\vspace{-2.2em}
\caption{Illustration of different featurizations. The curved triangle represents a micro surface patch in 3D with rich texture details, which can be reflected by the close-up imagery of a real-world large scene. Let the bold purple dot represent the target pixel colour $\boldsymbol{c}$, and we denote by the black arrow $\boldsymbol{d}$ the view direction of the camera ray. (a) UV-based featurizations~\cite{10274871, thies2019deferred, liu2023real} tend to disorganize the feature distribution due to distortions~\cite{floater2005surface, Hormann2007MeshPT, Ray2006PeriodicGP} in surface parametrization $\psi$. (b) Existing 3D-surface-based featurizations~\cite{aliev2020neural, ruckert2022adop, zuo2022view, rakhimov2022npbg++, kopanas2022neural, yang2022neumesh, kerbl20233d} fail to express the sub-primitive-scale intricate details given the limited discretization resolution. (c) Volumetric featurizations~\cite{muller2022instant, li2023neuralangelo, rosu2023permutosdf, wang2023neus2, barron2023zip} inevitably yield a dispersed weight distribution during volume rendering, where many multi-view inconsistent yet highly weighted samples ambiguate surface colour and deteriorate surface features with inconsistent colour gradient. (d) Our method leverages hash encoding to unlock the dependence of featuremetric resolution on discretization resolution, and utilizes rasterization to fully unleash the expressivity of volumetric hash encoding by propagating clean and multi-view consistent signals to surface features. 
}
\vspace{-1.0em}
\label{fig:method}
\end{figure*}
\section{Methodology}
\label{sec:method}
In this section, we introduce hash featurized manifold, aiming at high-fidelity cross-scale NVS of real-world large-scale scenes. Given a set of posed multi-view images $\{\mathcal{I}_{k}\}$ and a mesh $\mathcal{S}$ reconstructed using off-the-shelf MVS techniques, we conduct an expressive and distortion-free surface-based featurization with multi-resolution hash encoding. We then propose a rasterization pipeline and a neural shader tailored to the novel featurization for efficient, robust and high quality rendering. 

In the following, we first review the basic building blocks of our method (Sec.~\ref{sec:analysis}), then formulate our representation (Sec.~\ref{sec:representation}) and introduce several specific designs (Sec.~\ref{sec:improvements}) to further enhance the rendering quality.

\subsection{Preliminaries}
\label{sec:analysis}
\par\noindent\textbf{Deferred Neural Rendering.} Approaches like~\cite{aliev2020neural, ruckert2022adop, zuo2022view, rakhimov2022npbg++, kopanas2022neural, 10274871, thies2019deferred, liu2023real, yang2022neumesh, kerbl20233d} exploit various featurizations upon the explicit 3D discretization of the scene (typically an imperfect MVS reconstruction), and seamlessly incorporate graphics rasterization pipeline to enable photo-realistic neural rendering.

Given an explicit surface proxy $\mathcal{S}$ in the form of the triangle mesh or point cloud, these methods augment $\mathcal{S}$ with UV-based featurization $\mathcal{F}_{uv}: \mathbb{R}^2 \mapsto \mathbb{R}^Z$ or 3D-surface-based featurization $\mathcal{F}_{s}: \mathbb{R}^3 \mapsto \mathbb{R}^Z$, where learnable $Z$-dimensional feature descriptors are assigned to each surface primitive, which can be the UV texel~\cite{10274871, thies2019deferred, liu2023real}, mesh vertex~\cite{yang2022neumesh}, 3D point~\cite{aliev2020neural, ruckert2022adop, zuo2022view, rakhimov2022npbg++, kopanas2022neural}, or Gaussian~\cite{kerbl20233d}. Taking 3D-surface-based methods~\cite{aliev2020neural, ruckert2022adop, zuo2022view, rakhimov2022npbg++, kopanas2022neural, kerbl20233d, yang2022neumesh} as an example, given a target camera pose for rendering, the rasterizer $\mathcal{R}_{s}$ assigns each pixel a 3D point $\boldsymbol{x}_s \in \mathcal{S}$ at which the ray traced from the pixel intersects with $\mathcal{S}$, and then efficiently samples the corresponding feature $\mathcal{F}_{s}(\boldsymbol{x}_s)$. This essentially results in a screen-space feature buffer $\mathcal{R}_{s}(\{\mathcal{F}_{s}(\boldsymbol{x}_s)\}) \in \mathbb{R}^{H\times W\times Z}$.  Finally, a decoder $\mathcal{M}$ is utilized to interpret the feature buffer to synthesize the final RGB rendering $\mathcal{I} \in \mathbb{R}^{H\times W\times 3}$, which {can be parametrized as} CNNs~\cite{aliev2020neural, ruckert2022adop, rakhimov2022npbg++, zuo2022view}, MLPs~\cite{kopanas2022neural}, or spherical harmonic (SH) composition~\cite{kerbl20233d}, optionally taking as input the view direction map $\{\boldsymbol{d}_i\} \in \mathbb{R}^{H\times W\times 3}$. Putting pieces together, the deferred neural rendering framework can be formulated as:
\begin{equation}\label{eqn:dnr}
    \mathcal{I} = \mathcal{M} ( \mathcal{R}_{s}(\{\mathcal{F}_{s}(\boldsymbol{x}_s)\}), \{\boldsymbol{d}_i\}).
\end{equation}

\par\noindent\textbf{Multi-resolution Hash Encoding.}
State-of-the-art neural field based methods~\cite{li2023neuralangelo, wang2023neus2, rosu2023permutosdf, barron2023zip, muller2022instant} leverage a hybrid representation, i.e., a combination of hash encoding~\cite{muller2022instant} and a shallow MLP-based decoder, to efficiently reconstruct the geometry and appearance. 

In practice, hash encoding is conducted in a multi-resolution manner to effectively handle the collision. Let $\mathcal{V}$ be the continuous volume to be featurized, we denote by $\mathcal{F}_{v}^{\mathcal{H}, (\ell)}: \mathbb{R}^{3} \mapsto \mathbb{R}^{Z}$ the $\ell$-th level volumetric hash encoding, 
which conceptually expresses a dense 3D feature grid 
by hashing a learnable array $\mathcal{H} \in \mathbb{R}^{Z}$ of fixed length $N_h$. Given a 3D point $\boldsymbol{x} \in \mathcal{V}$, the corresponding hash feature $\mathcal{F}_{v}^{\mathcal{H}, (\ell)}(\boldsymbol{x}) \in \mathbb{R}^{Z}$ is queried by tri-linearly interpolating the hash entries at the vertices of the grid cell encompassing $\boldsymbol{x}$. The hash features across all resolutions are concatenated as: $\mathcal{F}_{v}^{\mathcal{H}}(\boldsymbol{x}) = \big\{\mathcal{F}_{v}^{\mathcal{H}, (\ell)}(\boldsymbol{x})\big|_{\ell=1}^{L}\big\} \in \mathbb{R}^{LZ}$, which is then passed to the light-weight decoder to reason about the implicit field. 
\subsection{Hash Featurized Manifold}
\label{sec:representation}
\begin{figure*}[htbp]
    \centering
    \newcommand{\colw}{0.19}
    \newcommand{\figw}{1} 
    \includegraphics[width=\figw\textwidth,trim={0cm 0cm 0cm 0cm},clip]{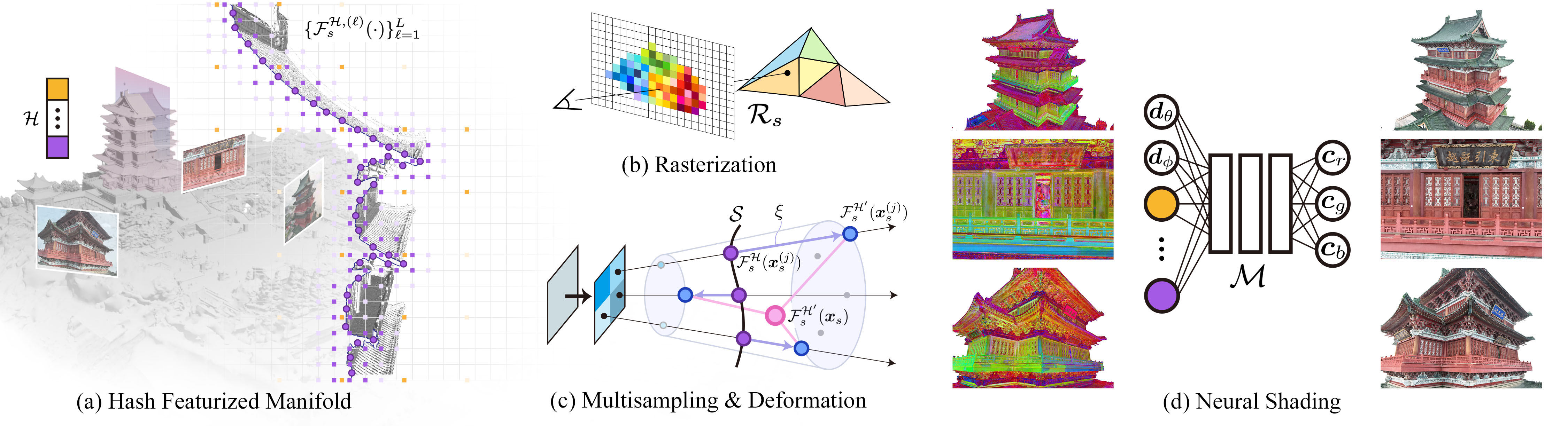}
    \hfill
\vspace{-2em}
\caption{An overview of the hash featurized manifold representation and our neural rendering framework. (a) We first reconstruct the scene as a mesh using MVS and featurize the surface manifold with volumetric multi-resolution hash encoding. (b) We then rasterize the featurized manifold into screen space and (c) optionally perform surface multisampling and manifold deformation to express a deformable frustum {for a better representation of the cross-scale details}. (d) An MLP-based neural shader decodes the rasterized feature buffer and account for the view dependent colour. Remarkably, we leverage rasterization to concentrate the featurization on multi-view consistency throughout the optimization, inherently converting the redundant volumetric featurization into an expressive surface-based featurization.}
\vspace{-1.0em}
\label{fig:pipeline}
\end{figure*}
Despite the notable advancements, existing scene representations become problematic in capturing the cross-scale, in-the-wild richness of large-scale scenes. In the following, we start by identifying several critical limitations of existing representations, 
which are illustrated in Fig.~\ref{fig:method} for better intuitions. 

For UV-based methods (Fig.~\ref{fig:method} (a)), 
the severely distorted parametrization, which is commonly encountered in large-scale scenarios with highly complicated shapes, essentially leads to a {disorganized feature distribution on the 3D surface $\mathcal{S}$ without preserving the conformality}, thus leading to stretched and blurry artifacts at local details. On the other hand, as shown in Fig.~\ref{fig:method} (b), existing representations based on explicit 3D discretization only 
{assign a single feature descriptor to each surface primitive}, thus failing to faithfully describe the intricate details {within the surface primitive} (e.g., the elliptic gaussians shaded in purple). For implicit volumetric representations (Fig.~\ref{fig:method} (c)), the weight distribution of volume rendering is scattered throughout the optimization, i.e., there exist many multi-view inconsistent yet highly weighted samples (e.g., the red and blue dots) that contaminate the supervision of surface colour and mislead the adaptations of surface features by propagating inaccurate colour gradient. 

To address the above limitations, we propose a novel scene representation, namely hash featurized manifold, aiming for the exploration of a more expressive surface-based featurization leveraging multi-resolution hash encoding and deferred neural rendering. 

We now give the detailed description of our design. 
Similar to existing neural representations based on explicit 3D discretization, we first reconstruct a mesh using off-the-shelf MVS techniques to serve as a 3D surface proxy of the scene. Then, we compute the bounding volume $\mathcal{V}$ of the mesh $\mathcal{S}$ and featurize it with volumetric multi-resolution hash encoding $\mathcal{F}_{v}^{\mathcal{H}, (\ell)}: \mathbb{R}^{3} \mapsto \mathbb{R}^{Z}$, which gives us a hash featurized volume $\{\mathcal{F}_{v}^{\mathcal{H}, (\ell)}(\boldsymbol{x})\}_{\boldsymbol{x} \in \mathcal{V}}$. Throughout the optimization, we leverage the mesh rasterizer $\mathcal{R}_s$ to calculate the 3D surface intersection $\boldsymbol{x}_s$ for each pixel and query the multi-resolution hash feature $\mathcal{F}_{v}^{\mathcal{H}, (\ell)}(\cdot) \in \mathbb{R}^{Z}$ only at the surface intersections $\boldsymbol{x}_s \in \mathcal{S}$ 
instead of in the redundant volume $\boldsymbol{x} \in \mathcal{V}$. With the explicit guidance of $\mathcal{S}$, the learnable hash table $\mathcal{H} \in \mathbb{R}^{Z}$ is forced to prioritize multi-view consistent surface regions $\{\boldsymbol{x} \in \mathcal{S} | \boldsymbol{x} \in \mathcal{V}\}$ with the most important fine scale features, inherently turning the redundant volumetric featurization into an expressive surface-based featurization:
\begin{equation}\label{eqn:hfm_analog}
    \{\mathcal{F}_{v}^{\mathcal{H}, (\ell)}(\boldsymbol{x})\}_{\boldsymbol{x} \in \mathcal{V}} \mapsto \{\mathcal{F}_{s}^{\mathcal{H}, (\ell)}(\boldsymbol{x}_s)\}_{\boldsymbol{x}_s \in \mathcal{S}}.
\end{equation}

Compared to vanilla hash-based volumetric featurization $\mathcal{F}_{v}^{\mathcal{H}, (\ell)}$ (Fig.~\ref{fig:method} (c)), our featurization $\mathcal{F}_{s}^{\mathcal{H}, (\ell)}$ (Fig.~\ref{fig:method} (d)) samples a single surface intersection along the pixel ray, eliminating the surface colour ambiguity. Moreover, our representation 
essentially concentrates on multi-view consistent surface and propagates clean, accurate gradient signals to surface features throughout the optimization, thus boosting the expressivity of multi-resolution hash encoding to faithfully describe the surface colour. Different from existing 3D-surface-based featurizations $\mathcal{F}_s$ (Fig.~\ref{fig:method} (b)), our featurization utilizes surface-aware hash encoding to effectively capture the sub-primitive-scale details regardless of the discretization resolution, demonstrating superior scalability towards large-scale scenes and cross-scale contents. In addition, our method allocates hash features on regular 3D voxel grids without relying on surface parametrizations, circumventing the distortion issues in UV-based featurizations $\mathcal{F}_{uv}$ (Fig.~\ref{fig:method} (a)). 

An overview of our deferred neural rendering pipeline is illustrated in Fig.~\ref{fig:pipeline}. Let $\mathcal{F}_{s}^{\mathcal{H}}(\boldsymbol{x}_s) = \big\{\mathcal{F}_{s}^{\mathcal{H}, (\ell)}(\boldsymbol{x}_s)\big|_{\ell=1}^{L}\big\} \in \mathbb{R}^{LZ}$ be the concatenation of hash features across all encoding resolutions, and {given a target camera pose}, we start by rasterizing the hash featurized manifold into screen space to create a hash-based feature buffer $\mathcal{R}_{s}(\{\mathcal{F}_{s}^{\mathcal{H}}(\boldsymbol{x}_s)\}) \in \mathbb{R}^{H\times W\times LZ}$. To enable realistic NVS, we use a light-weight MLP-based decoder $\mathcal{M}: \mathbb{R}^{LZ} \times \mathbb{R}^3 \mapsto \mathbb{R}^3$ to compute the view-dependent colour, {taking as inputs the resulting hash features and the view direction map $\{\boldsymbol{d}_i\}$}. The proposed neural rendering pipeline is therefore defined as:
\begin{equation}\label{eqn:hfm_dnr}
    \mathcal{I} = \mathcal{M}(\mathcal{R}_{s}(\{\mathcal{F}_{s}^{\mathcal{H}}(\boldsymbol{x}_s)\}), \{\boldsymbol{d}_i\}).
\end{equation}


\begin{figure*}[htbp]
    \newcommand{\colw}{0.5}
    \newcommand{\figw}{1.0} 
    \centering
    \includegraphics[width=\figw\textwidth,trim={0cm 0cm 0cm 0cm},clip]{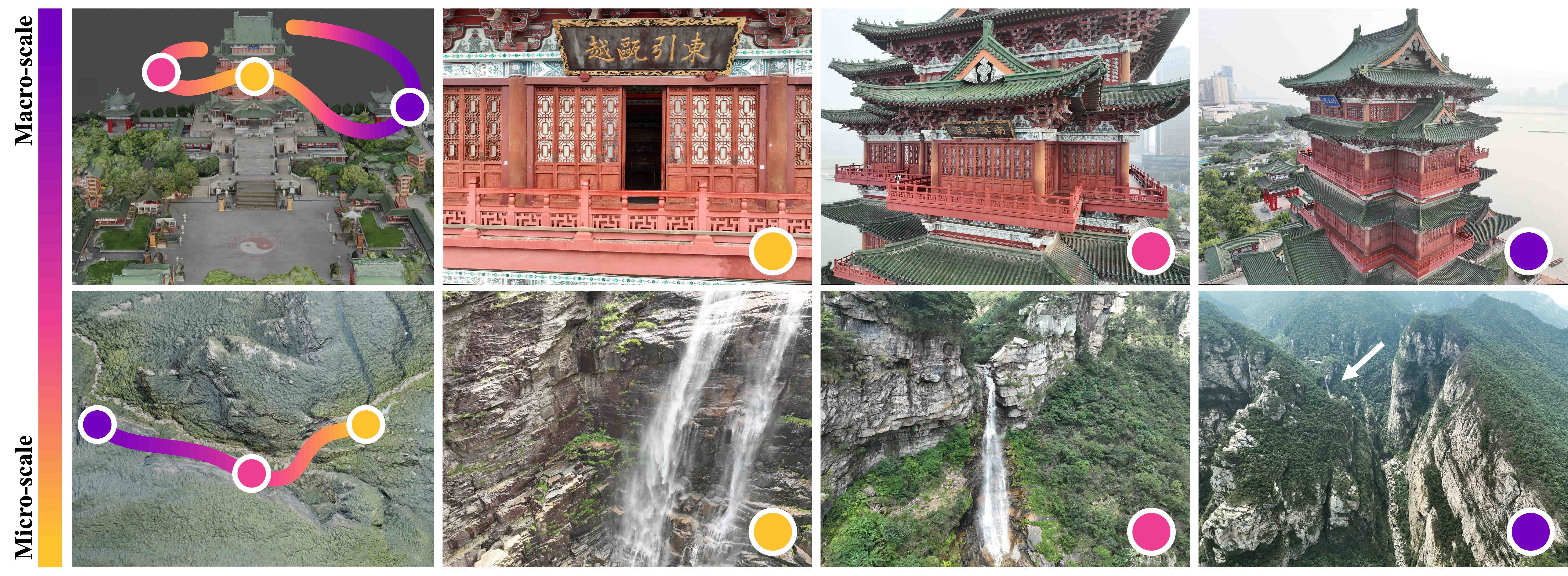}
    \vspace{-2.0em}
    \caption{Illustration of the real-captured, unstructured, cross-scale imagery in our \texttt{GigaNVS} dataset. We collect high quality multi-view images at varying distances ranging from $5m$ to $10^3m$.}
    \hfill
    \vspace{-1.5em}
    \label{fig:dataset}
\end{figure*}
\subsection{Enhancing Manifold Featurization}
\label{sec:improvements}
We have formulated the hash featurized manifold representation in Sec.~\ref{sec:representation} and this already delivers state-of-the-art NVS quality on challenging cross-scale scenarios. In this section, we further introduce two enhancements tailored for our representation, namely surface multisampling and manifold deformation, to better represent cross-scale details.

\par\noindent\textbf{Surface Multisampling.} Given the cross-scale, in-the-wild observations of a general large scene, casting a single ray per pixel neglects the unstructured scale variations and leads to blur or aliasing artifacts, due to the discrepancies in pixel colour when observing a surface point across varying distances or resolutions. 
To this end, we introduce a multisampling scheme for our surface-based featurization. Inspired by \cite{barron2023zip, greene1986creating, wang2022nerf}, we cast multiple rays per pixel 
to obtain a set of surface intersections $\{\boldsymbol{x}_{s}^{(j)}\}_{j=1}^{\gamma^2}$. To do so, we rasterize a $\gamma H \times \gamma W$ image, where each pixel in the original $H \times W$ image is super-sampled with a grid of $\gamma^2$ pixels. We then aggregate the information of the multiple surface intersections by individually querying the multi-resolution hash feature for each sample and pooling them with the mean operation:
\begin{equation}\label{eqn:mip}
    \mathcal{F}_{s}^{\mathcal{H}}(\boldsymbol{x}_s) = \sum_{j=1}^{\gamma^2}{\mathcal{F}_{s}^{\mathcal{H}}(\boldsymbol{x}_{s}^{(j)})} / \gamma^2.
\end{equation}
\par\noindent\textbf{Manifold Deformation.} Since our method directly featurizes a mesh, any geometric imperfections on the mesh will hinder the expressivity on local details. Similar to \cite{10274871}, we propose to further strengthen the multi-view consistency by a latent-space deformation $\xi: \mathbb{R}^{LZ} \times \mathbb{R}^{3} \mapsto \mathbb{R}^{LZ}$. Specifically, we first featurize the surface using another hash encoding $\mathcal{F}_{s}^{\mathcal{D}}(\cdot)$ with learnable hash table $\mathcal{D}$. Then, a tiny MLP $\xi$ takes as inputs the new hash features and the view direction vector $\boldsymbol{d} \in \mathbb{R}^3$ to deform the initial surface in high-dimensional feature space: 
\begin{equation}\label{eqn:disp}
    \mathcal{F}_{s}^{\mathcal{H}^{\prime}}(\boldsymbol{x}_{s}^{(j)}) = \mathcal{F}_{s}^{\mathcal{H}}(\boldsymbol{x}_{s}^{(j)}) + {\xi} \big(\mathcal{F}_{s}^{\mathcal{D}}(\boldsymbol{x}_{s}^{(j)}), \boldsymbol{d}\big).
\end{equation}

As shown in Fig.~\ref{fig:pipeline} (b), equipped with surface multisampling and manifold deformation, our hash featurized manifold essentially represents a deformable frustum near the initial surface, making it more robust to handle scale variations and more flexible at capturing micro-scale details. 

\section{Experiments}
\label{sec:exp}
\begin{figure*}[htbp]
    \centering
    \newcommand{\colw}{0.2}
    \newcommand{\figw}{1.0} 
    \includegraphics[width=\figw\textwidth,trim={0cm 0cm 0cm 0cm},clip]{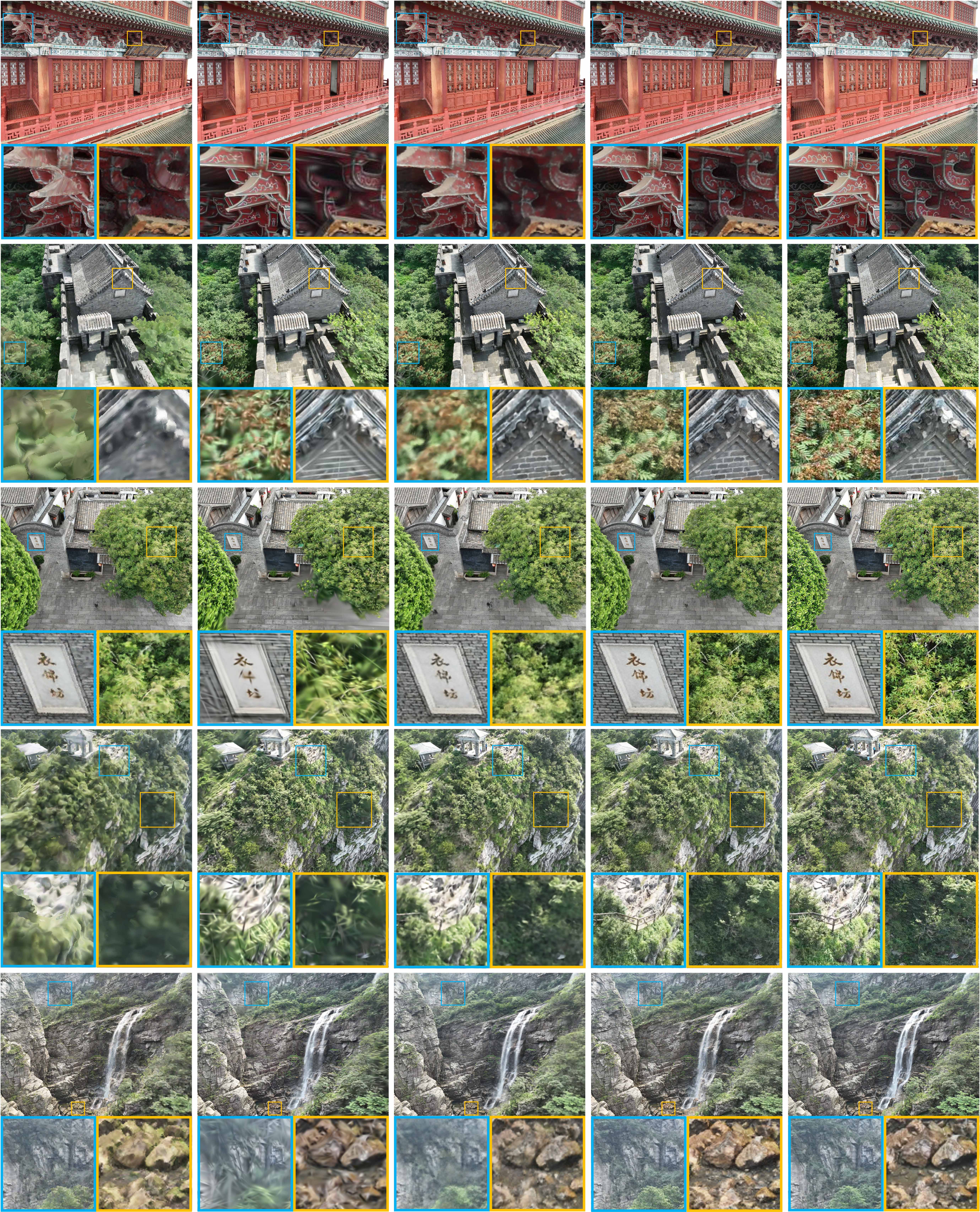}
\hfill
\vspace{-1.5em}
\centerline{\footnotesize{\hspace{-1.5em}
    (a) Meta Representation~\cite{10274871}\hspace{0.7em}(b) 3D Gaussian Splatting~\cite{kerbl20233d}\hspace{3.5em}(c) ZipNeRF~\cite{barron2023zip}\hspace{6.5em}(d) \textbf{Ours}\hspace{6.0em}
    (e) Ground Truth Image}}
\vspace{-1.5em}
\caption{Novel view synthesis results on the \texttt{GigaNVS} dataset. Compared to \cite{10274871, kerbl20233d, barron2023zip}, our method robustly synthesizes realistic colour and intricate details, preserving approximately the input-level resolution. Please zoom-in to see the details.}
\label{fig:qualitative result}
\end{figure*}


\par\noindent\textbf{Baselines.} We compare our method against explicit surface based representations~\cite{10274871, kerbl20233d} and implicit volumetric representations~\cite{muller2022instant, li2023neuralangelo} with the same MVS mesh as additional overhead. For 3DGS~\cite{kerbl20233d}, we initialize a dense set of gaussians on the mesh vertices and maintain as many gaussians as possible to make full use of the memory.  
For ZipNeRF~\cite{barron2023zip} and iNGP~\cite{muller2022instant}, we render the mesh into per-view depth map and supervise the volume-rendered depth similar to \cite{deng2022depth}. For Neuralangelo~\cite{li2023neuralangelo}, we directly supervise the 3D zero-level set by the mesh vertices as done in \cite{fu2022geo}. 
The hash encoding parameters in \cite{barron2023zip, muller2022instant, li2023neuralangelo} are set the same as ours. The implementation details can be found in the supplement.

\subsection{\textbf{\texttt{GigaNVS}} Dataset} 
We introduce a novel dataset, namely \texttt{GigaNVS}, to evaluate our method and the baselines on the challenging task, i.e. cross-scale NVS of real-world large-scale scenes. This dataset contains $7$ scenes of areas ranging from $1.3\times 10^{4}m^2$ to $3\times 10^{6} m^2$. For each scene, we collect thousands of high resolution (5K or 8K) multi-view images at multiple varying distances ($5m\sim10^3m$) using both aerial and ground photography. {The camera poses are calculated by Agisoft Metashape~\cite{agisoft} with subpixel-level reprojection errors}. We refer readers to the supplement for more details. 

An overview of our dataset is illustrated in Fig.~\ref{fig:dataset} and a comparison to existing real-world multi-view datasets is listed in Table~\ref{tab:giganvs}. To the best of our knowledge, \texttt{GigaNVS} is the first dataset characterized by gigantic scene scale and real-captured, cross-scale, high-resolution multi-view data. It fully exposes the scalability issues of state-of-the-art NVS algorithms, as verified in Sec.~\ref{sec:exp_results}, and conveys unprecedentedly rich contents of real-world large scenes, which are rarely valued in existing datasets. 

\begin{table}[htbp]
\scriptsize
\setlength{\tabcolsep}{0.7mm}{
\footnotesize
\begin{tabular}{cccccc}
\hline
\multicolumn{1}{c}{Datasets} & \#Scenes & \#Images & Area ($m^2$) & Resol. & Cross-scale \\ \hline
Tanks\&Temples~\cite{knapitsch2017tanks} & 21 &{$\sim400$} & {$\sim10^2$} & {2K} & {$\times$} \\ 
ETH3D~\cite{schops2017multi} & 25 & {$\sim40$} & {$\sim10^2$} & {6K} & {$\times$} \\
GigaMVS~\cite{zhang2021gigamvs} & 13 & {$\sim300$} & {$\sim10^4$} & {19K} & {$\times$} \\
MipNeRF360~\cite{barron2022mip} & 7 & {$\sim200$} & {$\sim10^1$} & {3K/5K} & {$\times$} \\
\textbf{\texttt{GigaNVS} (Ours)} & 7 & {$\sim7000$} & {$\sim10^6$} & {5K/8K} & {$\checkmark$} \\ 
\hline
\end{tabular}
}
\vspace{-1em}
\caption{Comparison of real-world multi-view datasets, where \#Images and Area denote the mean values across all scenes.}
\vspace{-2em}
\label{tab:giganvs}
\end{table}
\begin{table*}[htb]
\centering
\newcommand{\tabincell}[2]{\begin{tabular}{@{}#1@{}}#2\end{tabular}}
\footnotesize
\setlength{\tabcolsep}{0.4mm}{
\begin{tabular}{c|ccc|ccc|ccc|ccc|ccc}
\hline
\multirow{2}{*}{\tabincell{c}{Scene}} & \multicolumn{3}{c|}{Meta Representation~\cite{10274871}} & \multicolumn{3}{c|}{3DGS~\cite{kerbl20233d}} & \multicolumn{3}{c|}{Neuralangelo~\cite{li2023neuralangelo}} & \multicolumn{3}{c|}{ZipNeRF~\cite{barron2023zip}} & \multicolumn{3}{c}{\textbf{Ours}} \\ \cline{2-16}
{} & PSNR $\uparrow$ & SSIM $\uparrow$ & LPIPS$\downarrow$ & PSNR $\uparrow$ & SSIM $\uparrow$ & LPIPS$\downarrow$ & PSNR $\uparrow$ & SSIM $\uparrow$ & LPIPS$\downarrow$ & PSNR $\uparrow$ & SSIM $\uparrow$ & LPIPS$\downarrow$ & PSNR $\uparrow$ & SSIM $\uparrow$ & LPIPS$\downarrow$\\ \hline
TW-Pavilion (Day) & 20.63 & 0.668 & \cellcolor{Second}0.220 & \cellcolor{Second}21.52 & \cellcolor{Second}0.752 & \cellcolor{Third}0.225 & {18.31} & {0.553} & {0.394} & \cellcolor{Third}{21.26} & \cellcolor{Third}{0.691} & {0.260} & \cellcolor{First}{23.02} & \cellcolor{First}{0.786} & \cellcolor{First}{0.113} \\
TW-Pavilion (Night) & 23.82 & 0.750 & 0.235 & \cellcolor{Third}24.38 & \cellcolor{Second}0.816 & \cellcolor{Third}0.199 & {21.73} & {0.671} & {0.374} & \cellcolor{Second}{24.59} & \cellcolor{Third}{0.800} & \cellcolor{Second}{0.198} & \cellcolor{First}{25.55} & \cellcolor{First}{0.845} & \cellcolor{First}{0.120} \\
Lanes \& Alleys & 19.39 & 0.734 & \cellcolor{Third}0.188 & \cellcolor{Third}20.30 & \cellcolor{Third}0.787 & 0.222 & {18.10} & {0.633} & {0.321} & \cellcolor{Second}{20.39} & \cellcolor{Second}{0.796} & \cellcolor{Second}{0.169} & \cellcolor{First}{20.41} & \cellcolor{First}{0.814} & \cellcolor{First}{0.110} \\
The Great Wall (T3) & {16.94} & {0.458} & {0.439} & \cellcolor{Third}{18.11} & \cellcolor{Third}0.642 & \cellcolor{Third}0.389 & {14.25} & {0.196} & {0.731} & \cellcolor{Second}{18.19} & \cellcolor{Second}{0.651} & \cellcolor{Second}{0.325} & \cellcolor{First}{18.23} & \cellcolor{First}{0.685} & \cellcolor{First}{0.211} \\
The Great Wall (T2) & \cellcolor{Third}{19.86} & {0.701} & \cellcolor{Third}{0.226} & \cellcolor{Second}19.91 & \cellcolor{Third}0.709 & 0.314 & {19.01} & {0.635} & {0.319} & {18.99} & \cellcolor{Second}{0.780} & \cellcolor{Second}{0.202} & \cellcolor{First}{21.03} & \cellcolor{First}{0.806} & \cellcolor{First}{0.131} \\
The Five Old Peaks & \cellcolor{Third}18.98 & 0.614 & 0.330 & 18.79 & \cellcolor{Second}0.750 & \cellcolor{Third}0.272 & {16.65} & {0.417} & {0.555} & \cellcolor{Second}{19.93} & \cellcolor{Third}{0.741} & \cellcolor{Second}{0.251} & \cellcolor{First}{20.05} & \cellcolor{First}{0.774} & \cellcolor{First}{0.146} \\
{Sandie Spring} & {16.64} & {0.546} & {0.409} & \cellcolor{Third}17.25 & \cellcolor{Third}0.670 & \cellcolor{Third}0.380 & {13.98} & {0.264} & {0.678} & \cellcolor{Second}{17.32} & \cellcolor{Second}{0.679} & \cellcolor{Second}{0.298} & \cellcolor{First}{17.61} & \cellcolor{First}{0.724} & \cellcolor{First}{0.206} \\ \hline
Mean & 19.47 & 0.639 & 0.292 & \cellcolor{Third}20.04 & \cellcolor{Third}0.732 & \cellcolor{Third}0.286 & {17.43} & {0.481} & {0.482} & \cellcolor{Second}20.09 & \cellcolor{Second}0.734 & \cellcolor{Second}0.243 & \cellcolor{First}{20.84} & \cellcolor{First}{0.776} & \cellcolor{First}{0.148} \\ \hline
\end{tabular}
}
\vspace{-1em}
\caption{Quantitative comparisons on the \texttt{GigaNVS} dataset. Our method outperforms state-of-the-art approaches on all evaluation metrics.}
\label{tab:quantitative}
\vspace{-1.0em}
\end{table*}

\subsection{Comparative Results}
\label{sec:exp_results}
\par\noindent\textbf{Benchmark on \texttt{GigaNVS} Dataset\footnote[2]{Since the baselines can not directly consume high resolution (5K/8K) inputs due to memory issues, we perform fair comparisons using down-sampled images (1K) throughout the experiments. Please refer to the supplement for the high-resolution rendering of our method.}.}
In Table~\ref{tab:quantitative} we report mean PSNR, SSIM~\cite{wang2004image}, and LPIPS~\cite{zhang2018unreasonable} metrics across the test views in \texttt{GigaNVS} dataset. Our method outperforms all state-of-the-art neural rendering methods by a large margin, and remarkably, achieves a 40\% reduction in LPIPS relative to the second best method, ZipNeRF~\cite{barron2023zip}, demonstrating the significantly better perceptual fidelity of our method. The qualitative comparisons are shown in Fig.~\ref{fig:qualitative result}. Note that Meta representation~\cite{10274871} often produces blur and stretched artifacts due to the distortions of the UV-based featurization. 3DGS~\cite{kerbl20233d} delivers realistic rendering only at a global scale yet fails to capture the intricate details in close-ups. Implicit volumetric representations~\cite{barron2023zip, muller2022instant, li2023neuralangelo} suffer from surface ambiguities and struggle to handle complex large-scale structures even with geometric supervision, resulting in excessive blurries. By contrast, our method fully exploits the richness of the original imagery, enabling highly detailed NVS that is nearly indistinguishable from the ground truth. 

\par\noindent\textbf{Benchmark on Tanks\&Temples Dataset.} We compare against the state-of-the-arts on seven scenes from the Tanks\&Temples dataset~\cite{knapitsch2017tanks}, and we report the mean metrics across all selected scenes in Table~\ref{tab:small_tanks}. As observed, our method also demonstrates superiority on small-scale scenes from the public benchmark.
Please refer to the supplement for more detailed metrics reporting and visual comparisons.

\begin{table}[htbp]
\scriptsize
\setlength{\tabcolsep}{3.3mm}{
\footnotesize
\begin{tabular}{cccc}
\hline
\multicolumn{1}{c}{Methods} & PSNR$\uparrow$ & SSIM$\uparrow$ & LPIPS$\downarrow$ \\ \hline
Meta Representation~\cite{10274871} & {28.32} & {0.892} & {0.119} \\ 
3DGS~\cite{kerbl20233d} & \cellcolor{Third}{28.62} & \cellcolor{Third}{0.903} & {0.123} \\
Neuralangelo~\cite{li2023neuralangelo} & {27.41} & {0.898} & \cellcolor{Third}{0.114} \\
iNGP~\cite{muller2022instant} & \cellcolor{Second}{28.67} & \cellcolor{Second}{0.905} & \cellcolor{Second}{0.110} \\
\textbf{Ours} & \cellcolor{First}{29.66} & \cellcolor{First}{0.914} & \cellcolor{First}{0.079} \\
\hline
\end{tabular}
}
\vspace{-1em}
\caption{Quantitative evaluations on the Tanks\&Temples dataset. 
}\label{tab:small_tanks}
\end{table}
\begin{table}[htbp]
\scriptsize
\setlength{\tabcolsep}{4.2mm}{
\footnotesize
\begin{tabular}{cccc}
\hline
\multicolumn{1}{c}{Methods} & PSNR$\uparrow$ & SSIM$\uparrow$ & LPIPS$\downarrow$ \\ \hline
w/o Deformation & {22.77} & {0.775} & {0.131} \\ 
w/o Multisampling & {22.24} & {0.752} & {0.140} \\
Full Implementation & {23.02} & {0.786} & {0.113} \\ 
\hline
\end{tabular}
}
\vspace{-1em}
\caption{Ablation on featurization enhancements. 
}\label{tab:ablation}
\end{table}


\begin{table}[htbp]
\scriptsize
\setlength{\tabcolsep}{3.0mm}{
\footnotesize
\begin{tabular}{ccccc}
\hline
\multicolumn{1}{c}{Method} & \#Primitives & PSNR$\uparrow$ & SSIM$\uparrow$ & LPIPS$\downarrow$ \\ \hline
Ours & 10M & {23.02} & {0.786} & {0.113} \\ 
Ours & 1M & {22.84} & {0.782} & {0.114} \\
3DGS~\cite{kerbl20233d} & 3M & {21.52} & {0.752} & {0.225} \\
\hline
\end{tabular}
}
\vspace{-1em}
\caption{Ablation on mesh resolution. 
}\label{tab:ablation_mesh}
\end{table}
\begin{table}[htb]
\scriptsize
\setlength{\tabcolsep}{2.6mm}{
\begin{tabular}{cccc}
\cline{1-4}
\multicolumn{1}{c}{\multirow{2}{*}{Methods}} & 
\multicolumn{2}{c}{Time} &
\multicolumn{1}{c}{\multirow{2}{*}{GPU Mem.}} \\ \cline{2-3}
\multicolumn{1}{c}{} & {Optimization} & {Rendering} & \\ \cline{1-4}
Meta Representation~\cite{10274871} & 35 h  & \cellcolor{Third}0.34 s & {20 GB} \\
3DGS~\cite{kerbl20233d} & \cellcolor{Second}{1 h}  & \cellcolor{First}{0.01 s} & \cellcolor{Third}18 GB \\
ZipNeRF~\cite{barron2023zip} & {8.3 h}  & 14.3 s & {19 GB} \\
Neuralangelo~\cite{li2023neuralangelo} & 13 h  & 32.2 s & 19 GB \\
iNGP~\cite{muller2022instant} & \cellcolor{First}{0.8 h}  & 1.19 s & \cellcolor{First}{8 GB} \\
Ours & \cellcolor{Third}6.5 h & \cellcolor{Second}0.08 s & \cellcolor{Second}{15 GB} \\ \hline
\end{tabular}}
\vspace{-1em}
\caption{{Comparison of time and memory cost. 
}}\label{tab:table_efficiency}
\end{table}

\par\noindent\textbf{Ablations on Featurization Enhancements.} We ablate the surface multisampling and manifold deformation module in our pipeline using the TW-Pavilion (Day) scene. As shown in Table~\ref{tab:ablation}, both designs improve the rendering quality. 

\par\noindent\textbf{Ablations on Mesh Resolution.} In Table~\ref{tab:ablation_mesh}, we demonstrate the robustness of our method w.r.t. the mesh resolution, where mesh decimation~\cite{garland1997surface} is used to obtain the mesh with desired face counts. Notably, our method achieves less than 1\% of differences in all metrics when down-samping the mesh by 90\%, indicating the effectiveness of our featurization without heavy reliance on the 3D discretization resolution. Our representation with 1 million triangle faces significantly outperforms the state-of-the-art 3DGS~\cite{kerbl20233d} with 3 million gaussians. Metrics are reported on the TW-Pavilion (Day) scene, and we refer readers to the supplement for more ablations on other scenes.

\par\noindent\textbf{Efficiency.} We compare the time and memory cost of the competing methods in Table~\ref{tab:table_efficiency}, using the TW-Pavilion (Day) scene. The rendering time and GPU memory cost are measured by synthesizing an image of 989$\times$1320 resolution. Our method is significantly faster than volume rendering based methods~\cite{barron2023zip, muller2022instant, li2023neuralangelo} due to the rasterization pipeline. Compared to \cite{10274871}, our method also shows superior efficiency by using a light-weight neural shader. Note that 3DGS~\cite{kerbl20233d} demonstrates the highest time efficiency leveraging the splatting and SH-based calculations, whereas iNGP~\cite{muller2022instant} and ours are more efficient in memory by incorporating compact neural components. 


\section{Conclusion}
\label{sec:conclusion}
We introduce hash featurized manifold, a novel representation for high-fidelity cross-scale neural rendering of real-world large-scale scenes. The core is an expressive surface-based featurization constructed by guiding the volumetric hash encoding with the rasterization of a surface manifold. Our representation fully unlocks the expressivity of multi-resolution hash encoding by skipping the redundant space and concentrating on multi-view consistent colour gradient. 
We also propose a novel \texttt{GigaNVS} dataset consisting of seven real-world large scenes to benchmark cross-scale, high-resolution novel view synthesis. Extensive experiments demonstrate that our method achieves unparalleled levels of realism by effectively reflecting both macro-scale and micro-scale scene contents.

\noindent\textbf{Limitation \& Future Work}
Although showing robustness to the mesh resolution, our method currently can not handle the incompleteness and occlusions caused by the incorrect geometry. Our future work is to exploit differentiable rendering for more flexible control of the geometry.

\noindent\textbf{Acknowledgements} 
This work is supported in part by Natural Science Foundation of China (NSFC) under contract No.~62125106, 61860206003, 62088102, 62171256, 62331006, in part by National Science and Technology Major Project under contract No.~2021ZD0109901.\\
\vspace{-3em}
{
    \small
    \bibliographystyle{ieeenat_fullname}
    \bibliography{main}
}


\end{document}